\documentclass[10pt,letterpaper]{article}

\usepackage[T1]{fontenc}
\usepackage[utf8]{inputenc}
\usepackage[margin=1in]{geometry}
\usepackage{mathpazo}
\usepackage{inconsolata}
\usepackage{microtype}
\usepackage{amsmath,amssymb}
\usepackage{booktabs}
\usepackage{array}
\usepackage{graphicx}
\usepackage{xcolor}
\usepackage{enumitem}
\usepackage{float}
\usepackage[ruled,vlined,linesnumbered]{algorithm2e}
\usepackage[authoryear,round]{natbib}
\setcitestyle{citesep={;},aysep={,},yysep={;}}
\usepackage{url}
\usepackage{hyperref}

\definecolor{darkblue}{rgb}{0,0,0.5}
\hypersetup{
  colorlinks=true,
  citecolor=darkblue,
  linkcolor=darkblue,
  urlcolor=darkblue,
  pdftitle={AvoKV-E: Payload-Aware KV Cache Eviction for Long Reasoning},
  pdfauthor={Han Yu; Wenhui Zhu; Xiwen Chen; Zhipeng Wang; Hejian Sang; Jincheng Cao; Yuanda Xu; Zhengze Zhou; Han Shi; Menglin Zhou; Xuanzhao Dong; Minzhou Huang; Rui Cai; Hao Wang; Alborz Geramifard}
}

\newcommand{\method}{AvoKV-E}
\newcommand{\rkv}{R-KV}
\newcommand{\thinkv}{ThinKV}
\newcommand{\lazy}{LazyEviction}
\newcommand{\fullkv}{FullKV}

\title{\bfseries \method{}: Payload-Aware KV Cache Eviction for Long Reasoning}
\author{%
  \normalsize
  \begin{tabular}{c}
    Han Yu$^{1}$, Wenhui Zhu$^{1}$, Xiwen Chen$^{2}$, Zhipeng Wang$^{1}$, Hejian Sang$^{1}$ \\
    Jincheng Cao$^{1}$, Yuanda Xu$^{1}$, Zhengze Zhou$^{1}$, Han Shi$^{3}$, Menglin Zhou$^{1}$ \\
    Xuanzhao Dong$^{4}$, Minzhou Huang$^{1}$, Rui Cai$^{1}$, Hao Wang$^{2}$, Alborz Geramifard$^{1}$ \\[0.6em]
    $^{1}$LinkedIn Corporation \quad $^{2}$Clemson University \\
    $^{3}$Rice University \quad $^{4}$Arizona State University \\[0.4em]
    \texttt{hanyu@linkedin.com}
  \end{tabular}%
}
\date{}

\begin{document}
\maketitle

\begin{abstract}
Long-output reasoning shifts the KV-cache bottleneck from the fixed prompt to the generated trace. Existing reasoning-cache eviction methods largely treat cached entries as routing objects, estimating whether an old key will still be read, will recur, or can be replaced. This routing-only view overlooks two effects: low-attention entries can carry large value payloads whose removal changes future predictions, and newly generated states can appear stale before later queries have had a chance to read them. We introduce \method{}, a training-free eviction policy that first delays eligibility for recent states and then ranks eligible entries using candidate-normalized read pressure, key redundancy, and value-payload potential. According to empirical evaluation across different models and datasets, \method{} matches or exceeds redundancy-aware, recurrence-based, and thought-adaptive eviction baselines at matched active-KV budgets, with its largest gains in the tightest-cache regime. Component and counterfactual analyses further connect these gains to delayed observation, payload-aware scoring, redundancy, and scale-robust normalization. Together, the results show that long-reasoning KV eviction should preserve not only keys that are likely to be read, but also the value payloads that sustain the reasoning trajectory.
\end{abstract}

\section{Introduction}
\label{sec:intro}

\begin{figure*}[t]
\centering
\includegraphics[width=0.98\textwidth]{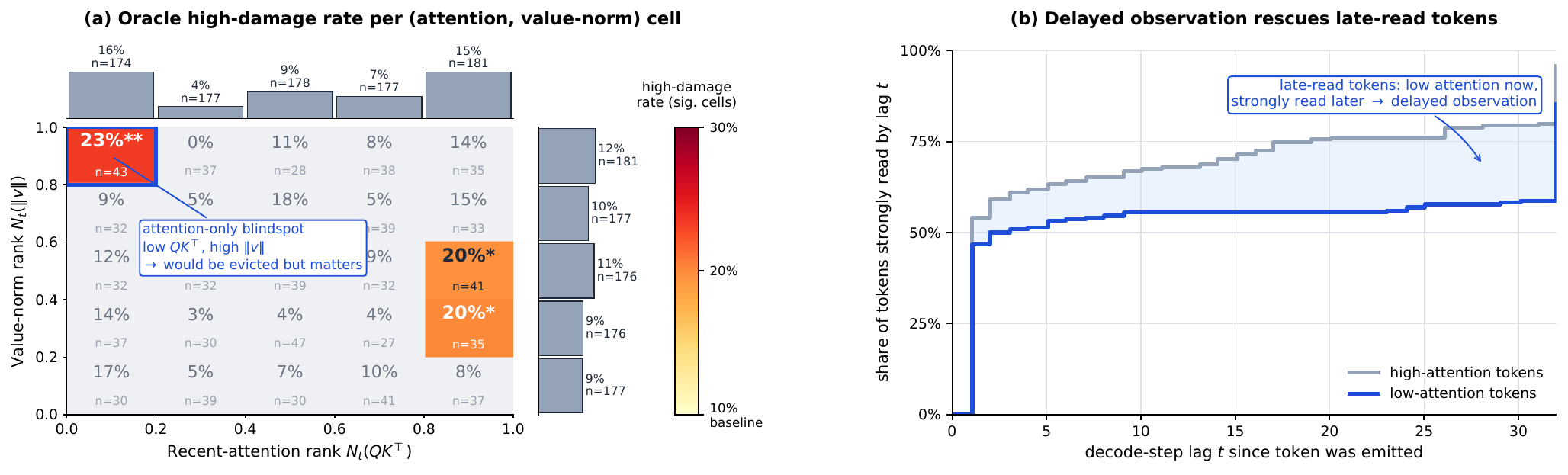}
\caption{Diagnostic evidence for two failure modes of routing-only eviction. \textbf{(a)}~Some low-attention entries with large value payloads are nevertheless high-damage to remove, revealing an \emph{attention-only blindspot}. \textbf{(b)}~Some low-attention entries receive strong reads only after the initial observation window, so early eviction can confuse ``not yet read'' with ``not useful.'' Detailed construction, counts, and statistical tests are given in Appendix~\ref{app:motivation}.}
\label{fig:motivation}
\end{figure*}

Large reasoning models such as DeepSeek-R1 \citep{deepseekr12025} and Qwen3 \citep{qwen3technicalreport2025} often solve difficult problems by producing long derivations, intermediate calculations, and self-corrections. Every generated token appends a key and a value at every layer, turning the reasoning trace into a growing memory object. Under a tight active-cache budget, an eviction mistake can therefore disrupt the reasoning process itself: the model may revisit completed work, enter a loop, or fail to produce a final answer.

Most KV-cache policies decide what to retain by asking whether an entry is likely to be retrieved again. Attention histories, recurrence, redundancy, and thought-level structure provide different signals for answering this question \citep{scissorhands2023,tova2024,h2o2023,streamingllm2023,snapkv2024,nacl2024,rkv2025,lazyeviction2025,thinkv2025,longflow2026,raas2025}. At a high level, however, these methods primarily evaluate the cache through its keys: will an entry receive future attention, or can a nearby key serve as a substitute?

This routing-oriented view misses two properties that matter in long reasoning. First, a KV entry has two distinct roles: its key acts as a retrieval address, while its value contains the vector returned when that address is read. Eviction removes both, so an entry that appears unimportant from its attention pattern can still carry consequential content. Second, a newly generated state may appear unimportant simply because later queries have not yet had an opportunity to use it. A policy that scores such a state immediately can confuse ``not yet read'' with ``not useful.''

Figure~\ref{fig:motivation} illustrates both effects in decoded MATH/AIME traces. A leave-one-out replay diagnostic shows that the low-attention/high-payload region is enriched by $2.4{\times}$ in entries whose removal substantially changes future predictions. A complementary lag diagnostic shows that some low-attention states receive strong reads only after the initial observation window. Together, these observations suggest that early attention alone provides incomplete evidence for eviction: it can overlook what an entry carries and judge the entry before evidence of its future use has arrived. Appendix~\ref{app:motivation} gives the full diagnostic construction, candidate counts, and statistical tests.

These observations motivate \method{}, a two-stage, training-free eviction policy that separates \emph{when} an entry may be judged from \emph{how} it should be scored. First, delayed causal observation protects newly generated states until subsequent queries have had an opportunity to reveal their utility. The method then ranks eligible entries using three complementary signals: read pressure measures continued retrieval, key redundancy measures whether a nearby substitute exists, and value-payload potential measures the content that would be lost upon eviction. Candidate-relative normalization places these signals on a comparable scale throughout decoding.

We evaluate \method{} on MATH-500 and AIME-25 using DeepSeek-R1-Distill-Llama-8B and Qwen3-14B. At matched active-KV budgets, it matches or outperforms redundancy-aware, recurrence-based, and thought-adaptive baselines, with its largest gains in the tightest-cache regime. Component ablations and counterfactual oracle analysis further connect the end-task improvements to delayed observation, payload-aware scoring, redundancy, and candidate-relative normalization.

\begin{samepage}
\paragraph{Contributions.} Our contributions are three-fold:
\begin{itemize}[leftmargin=*,nosep]
  \item We identify two limitations of routing-only eviction in long reasoning: it can overlook consequential value payloads and evaluate recent states before their utility becomes observable.
  \item We introduce \method{}, a training-free policy that combines delayed eligibility with candidate-normalized read pressure, key redundancy, and value-payload potential.
  \item We evaluate \method{} across two reasoning models and two mathematical reasoning benchmarks, and analyze its components, oracle ranking quality, and termination behavior.
\end{itemize}
\end{samepage}

\section{Related Work}

\textbf{Reasoning-cache compression.} Long-output reasoning makes the generated trace itself a dominant KV-cache object. Recent methods adapt compression to this setting using reasoning milestones or current-query importance \citep{raas2025,longflow2026}, key redundancy \citep{rkv2025}, lagged recurrence of attention importance \citep{lazyeviction2025}, or thought-adaptive compression across layers \citep{thinkv2025}. As an online eviction policy for generated reasoning states, \method{} is closest to \rkv{} and \lazy{}. \rkv{} asks whether a key is important and non-redundant, while \lazy{} asks whether attention importance recurs after a lag. In contrast, the delay in \method{} determines when an entry becomes eligible rather than acting as an importance score; once eligible, the entry is evaluated for the risk of discarding both its routing address and its value content. The method therefore requires neither a reasoning-milestone detector nor a thought-regime classifier.

\textbf{Attention-based retention and observation.} A broader line of KV-cache compression estimates future utility from attention history. Scissorhands exploits the persistence of token importance, H2O preserves attention heavy hitters, StreamingLLM protects attention sinks, and TOVA updates the cache using recent attention \citep{scissorhands2023,h2o2023,streamingllm2023,tova2024}. SnapKV and NACL instead aggregate statistics over a short observation window before selecting retained entries \citep{snapkv2024,nacl2024}. Although their retention rules differ, these methods primarily evaluate an entry as an attention destination. CAOTE moves closer to the consequence of eviction by estimating attention-output error \citep{caote2025}. \method{} shares this concern, but uses an online value-norm proxy inside a reasoning-cache loop and delays the scoring of recent states whose future use has not yet become observable.

\textbf{Budget allocation and query-aware access.} A complementary family decides where limited capacity should be spent across layers, heads, or queries \citep{fastgen2024,pyramidkv2024,adakv2024,headkv2024,quest2024,duoattention2024}. These methods change the allocation or access structure of the cache rather than only the score used to rank individual entries under a fixed budget. Our study holds allocation uniform across layers and heads so that it can isolate the token-level eviction decision. Within this controlled setting, \method{} asks which generated reasoning states should remain once the available capacity has been specified.

\textbf{Quantization and serving systems.} KV-cache quantization reduces the memory occupied by retained states through lower or adaptive precision \citep{kivi2024,kvquant2024,pmkvq2025,turboquant2026}, while PagedAttention improves how KV memory is managed by the serving system \citep{vllm2023}. These techniques address storage precision and memory organization, respectively, and are complementary to the eviction policy studied here. We retain full-precision entries and use a shared decoding harness so that the experiments compare which reasoning states are kept rather than differences in quantization or serving infrastructure.

\section{Problem Formulation}
\label{sec:problem}

\textbf{Causal KV-cache compression.}
Consider an autoregressive decoder at compression checkpoint $t$. Let $\mathcal{I}_t$ denote the indices of the KV entries currently stored in the active cache. Each entry $i\in\mathcal{I}_t$ contains a key $k_i^{\ell,h}$ and value $v_i^{\ell,h}$ at layer $\ell$ and head $h$. Given an active KV budget $B$, a compression policy selects a keep set
\begin{equation}
    K_t \subseteq \mathcal{I}_t,
    \qquad
    |K_t| \leq B.
    \label{eq:budget-constraint}
\end{equation}
Only entries in $K_t$ remain available to subsequent decoding. The selection must be causal: it may use the cached keys and values and attention patterns observed up to checkpoint $t$, but it cannot inspect future queries or their attention patterns.

The utility of a cached entry depends on how it will affect future predictions, yet this evidence is incomplete when the compression decision is made. In particular, a newly generated entry with low observed attention may be genuinely expendable, or it may simply be too recent for later queries to have used it. Moreover, each entry serves two roles: its key determines how it is retrieved, while its value determines what content is returned. Eviction therefore removes both a routing address and its associated payload. These ambiguities require the policy to answer two separate questions: \emph{when} has an entry accumulated enough causal evidence to be judged, and \emph{how} should eligible entries be ranked under the budget? Section~\ref{sec:method} describes how \method{} addresses these questions through delayed eligibility and payload-aware eviction risk.

Table~\ref{tab:notation} summarizes the notation used in the problem formulation, method, and analysis.

\begin{table}[t]
\centering
\small
\setlength{\tabcolsep}{3pt}
\begin{tabular}{ll}
\toprule
Symbol & Description \\
\midrule
$t$            & current decode step / compression checkpoint \\
$\mathcal{I}_t$ & active cache entries before compression \\
$L,H$          & number of layers and attention heads \\
$k_i^{\ell,h}, v_i^{\ell,h}$ & key/value of entry $i$ at layer $\ell$, head $h$ \\
$a_{\tau,i}^{\ell,h}$ & attention from query $\tau$ to entry $i$ \\
$W_{\mathrm{obs}}$  & observation-window length \\
$W_{\mathrm{delay}}$ & protected-delay length, $\geq W_{\mathrm{obs}}$ \\
$O_t$           & query positions used for scoring \\
$P_t$           & protected delay-buffer entries \\
$C_t$           & old eligible candidate entries \\
$\mathcal{N}_i$ & $C_t\setminus\{i\}$, the other redundancy candidates \\
$p_i, r_i, b_i$ & read pressure, redundancy, payload potential \\
$N_t(\cdot)$    & within-candidate percentile-rank normalizer \\
$S_i(t)$        & eviction-risk score of entry $i$ at $t$ \\
$\lambda_p, \lambda_r, \lambda_b$ & score weights \\
$B$             & active KV budget \\
$K_t$           & keep set after compression \\
$D_i^{\mathrm{KL}}$ & oracle leave-one-out KL of entry $i$ \\
\bottomrule
\end{tabular}
\caption{Notation used in the problem formulation, method, and analysis.}
\label{tab:notation}
\end{table}

\section{Method}\label{sec:method}

\begin{figure}[H]
\centering
\includegraphics[width=0.98\textwidth]{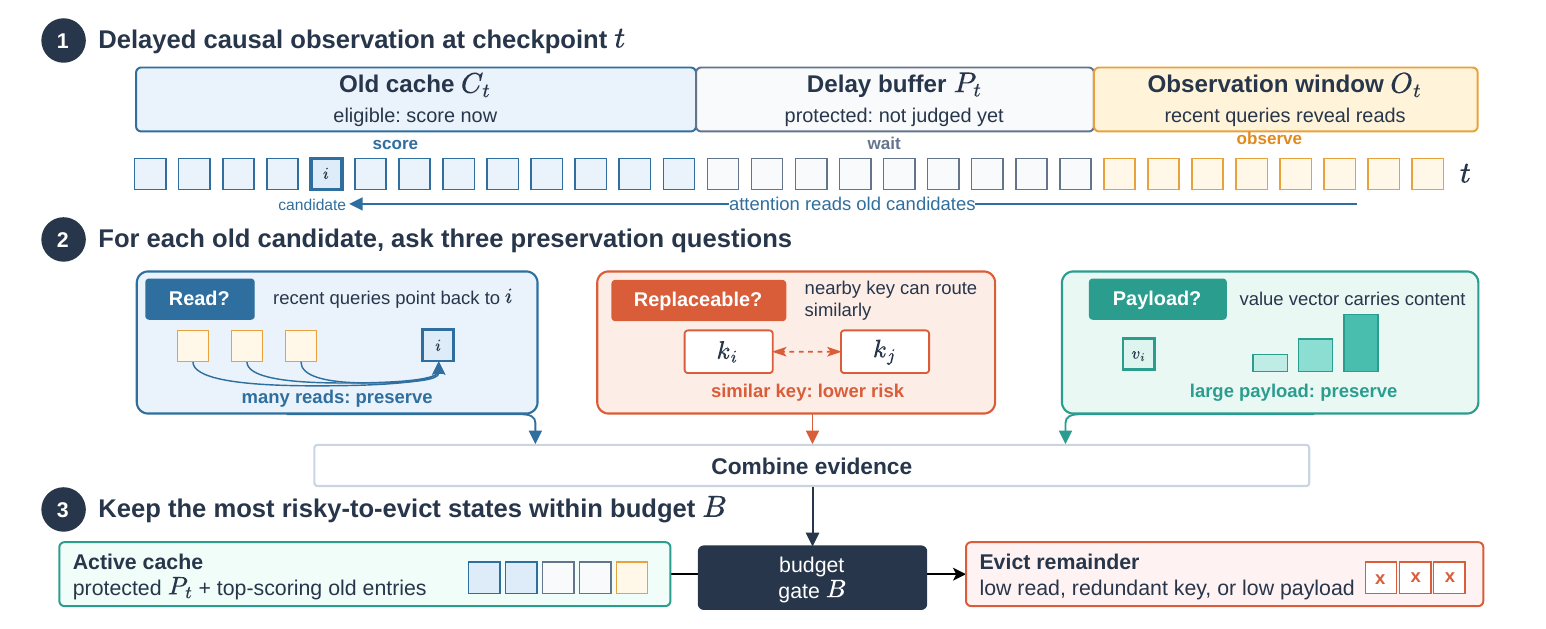}
\caption{Overview of \method{}. Recent entries in the delay buffer $P_t$ are temporarily protected. Older eligible candidates in $C_t$ are ranked by a scale-robust score $S_i(t)$ that combines read pressure $p_i$, key redundancy $r_i$, and value-payload potential $b_i$. The top-scoring entries are retained under active budget $B$.}
\label{fig:method}
\end{figure}

\textbf{Overview.} \method{} resolves this ambiguity by separating \emph{when} an entry may be judged from \emph{how} it is ranked once eligible. First, \emph{delayed causal observation} temporarily protects recent entries, preventing a newly generated state from being evicted before later queries have had an opportunity to read it. Second, a \emph{payload-aware eviction-risk score} ranks older candidates using three complementary signals: read pressure~$p_i$, which measures whether later queries still read the entry; key redundancy~$r_i$, which measures whether a nearby key can substitute for it and is inspired by \citep{rkv2025}; and value-payload potential~$b_i$, which measures whether removal would discard a salient value vector. Figure~\ref{fig:method} illustrates these two stages.

\subsection{Delayed Causal Observation}
\paragraph{Motivation.} A newly generated state may receive little attention simply because only a few subsequent queries have been available to read it. Scoring such a state immediately would conflate ``not yet observed'' with ``unimportant.'' \method{} therefore delays the eviction decision until the state has remained in the cache long enough to be observed causally.

\paragraph{Cache partition.} At checkpoint $t$, let $W_{\mathrm{obs}}$ be the number of recent query positions used to measure read pressure, and let $W_{\mathrm{delay}}\geq W_{\mathrm{obs}}$ be the period during which a newly generated entry remains protected. We define
\begin{align}
  O_t &= \{\tau: t-W_{\mathrm{obs}} < \tau \leq t\}, \\
  P_t &= \{i: t-W_{\mathrm{delay}} < i \leq t\}, \\
  C_t &= \{i: i \leq t-W_{\mathrm{delay}}\}.
\end{align}
\paragraph{Roles of the three sets.} The observation set $O_t$ contains the recent query positions used for scoring. The protected set $P_t$ contains newly generated entries that cannot yet be evicted, whereas the candidate set $C_t$ contains older entries that are eligible for ranking. Only entries in $C_t$ compete for the remaining cache budget after $P_t$ has been reserved.

\paragraph{Effect of the delay.} Because $W_{\mathrm{delay}}\geq W_{\mathrm{obs}}$, every candidate in $C_t$ predates every query in $O_t$. An entry is therefore exposed to eviction only after the queries used to measure its read pressure have had a causal opportunity to attend to it. The delay does not itself predict importance, as a recurrence model does \citep{lazyeviction2025}; it is a measurement rule that controls eligibility. Once an entry leaves $P_t$, the payload-aware score determines whether a low-read state should nevertheless be retained because it carries a large value payload.

\subsection{Payload-Aware Eviction Risk}
\paragraph{Motivation.} Once an entry is old enough to be judged, \method{} must estimate how risky it would be to remove. Read pressure describes whether the entry is currently being used, and key redundancy describes whether another key can serve a similar routing role. Neither signal, however, directly reflects the value payload that would be lost if the entry were removed. We therefore combine routing, substitutability, and payload information in a single eviction-risk score.

\paragraph{Score components.} For $i\in C_t$, we average three signals over layers and heads: $a_{\tau,i}^{\ell,h}$ is attention from query $\tau$; $k_i^{\ell,h},v_i^{\ell,h}$ are cached key/value; and $\mathcal{N}_i=C_t\setminus\{i\}$.
\begin{align}
  p_i(t)
  &=
  \frac{1}{|O_t|LH}
  \!\!\sum_{\tau\in O_t}\!\sum_{\ell,h}
  a_{\tau,i}^{\ell,h},
  \label{eq:read-pressure}\\
  r_i(t)
  &=
  \frac{1}{LH}\!\sum_{\ell,h}\!\max_{j\in\mathcal{N}_i}
  \cos(k_i^{\ell,h},k_j^{\ell,h}),
  \label{eq:redundancy}\\
  b_i(t)
  &=
  \frac{1}{LH}\!\sum_{\ell,h}
  \frac{\|v_i^{\ell,h}\|_2}
       {\operatorname{median}_{j\in C_t}\|v_j^{\ell,h}\|_2+\epsilon}.
  \label{eq:payload}
\end{align}

\textbf{Interpreting the components.} Read pressure $p_i$ averages the attention received from queries in $O_t$ and is high when the current context continues to read the entry. Redundancy $r_i$ compares the entry with its candidate neighborhood $\mathcal{N}_i$ in key space and is high when a nearby substitute exists. Payload potential $b_i$ compares the value norm with the per-layer/head median over $C_t$, providing a candidate-relative proxy for how much value content the entry carries. The three components thus answer different questions: whether the entry is read, whether its key is replaceable, and whether its value payload is salient.

Crucially, \method{} does \emph{not} multiply $b_i$ by $p_i$. Such a product would suppress exactly the low-read/high-payload states identified in Figure~\ref{fig:motivation}\,(a). Keeping the payload term independent allows an unusually large value vector to rescue a candidate even when its recent attention is low. This term is related in spirit to CAOTE's attention-output error \citep{caote2025}, but instantiates that intuition as an online value-norm proxy within the delayed reasoning-cache loop.

\textbf{Candidate-relative normalization.} The three raw components cannot be combined directly because they have different numerical scales. We therefore express each component by its percentile rank among the eligible candidates at the current checkpoint:
\begin{equation}
  N_t(x_i)
  =
  \frac{\operatorname{rank}_{C_t}(x_i)-1}{|C_t|-1},
  \label{eq:normalizer}
\end{equation}

where $\operatorname{rank}_{C_t}$ is ascending and ties are averaged. The smallest candidate value therefore maps to $0$, the largest maps to $1$, and all three components are placed on the same $[0,1]$ scale. We then combine the normalized components into the eviction-risk score
\begin{equation}
  S_i(t)
  =
  \lambda_p N_t(p_i)
  -
  \lambda_r N_t(r_i)
  +
  \lambda_b N_t(b_i).
  \label{eq:avokv-score}
\end{equation}

The signs encode the role of each component. High read pressure and high value-payload potential make eviction riskier and therefore increase $S_i(t)$, whereas high key redundancy makes eviction safer and decreases it. An entry can thus be retained because it is actively read, difficult to replace, or carries an unusually salient payload.

The protected initial entries remain in $C_t$ but receive $S_i(t){=}+\infty$, ensuring that the top-score selection always retains them. We protect the first four cache entries in all experiments.

Given active budget $B$, \method{} first reserves the protected set $P_t$ and fills the remaining $B-|P_t|$ slots with the highest-scoring candidates: $K_t = P_t \cup \operatorname{Top}_{B-|P_t|}\{S_i(t):i\in C_t\}$. We use $\lambda_p{=}0.1$, $\lambda_r{=}0.9$, and $\lambda_b{=}0.05$ throughout the main experiments and ablate each component in \S\ref{sec:ablations}.

\subsection{Online Compression}
Algorithm~\ref{alg:avokv-loop} summarizes one complete eviction checkpoint. Ordinary autoregressive generation remains unchanged between checkpoints. When the active cache is within budget, \method{} returns it unchanged. Otherwise, the method partitions entries into $P_t$ and $C_t$, computes and normalizes the three signals for every eligible candidate, retains all protected entries, and uses $S_i(t)$ to fill the remaining cache slots. Delayed observation and payload-aware ranking therefore act as two consecutive decisions: the former determines which entries may be compared, and the latter determines which of those entries are safest to evict.

\begin{algorithm}[tbp]
\SetAlFnt{\small}
\SetAlCapFnt{\small}
\SetAlCapNameFnt{\small}
\DontPrintSemicolon
\KwIn{active budget $B$; windows $W_{\mathrm{obs}}, W_{\mathrm{delay}}$; protected initial entries;
weights $\lambda_p,\lambda_r,\lambda_b$; candidate cache up to step $t$.}
\KwOut{keep set $K_t$ under the active KV budget.}
$O_t \gets \{\tau : t-W_{\mathrm{obs}} < \tau \leq t\}$
   \tcp*{observation window}
$P_t \gets \{i : t-W_{\mathrm{delay}} < i \leq t\}$
   \tcp*{protected delay buffer}
$C_t \gets \{i : i \leq t-W_{\mathrm{delay}}\}$
   \tcp*{old eligible entries}
\If{$|P_t| + |C_t| \leq B$}{
  \Return the cache unchanged\;
}
\ForEach{$i \in C_t$}{
  $p_i \gets$ avg attention from $O_t$, averaged over layers/heads\;
  $r_i \gets$ avg cosine of $k_i$ to its $\mathcal{N}_i$-best key\;
  $b_i \gets$ avg $\|v_i\|_2$, normalized by the per-layer/head median over $C_t$\;
}
$\hat p_i, \hat r_i, \hat b_i \gets N_t(p_i),\, N_t(r_i),\, N_t(b_i)$
   \tcp*{percentile-rank normalize}
$S_i \gets \lambda_p\, \hat p_i \;-\; \lambda_r\, \hat r_i \;+\; \lambda_b\, \hat b_i$
   \tcp*{eviction-risk score}
$S_i \gets +\infty$ for protected initial entries\;
$K_t \gets P_t \,\cup\, \operatorname{Top}_{B-|P_t|}\{S_i : i \in C_t\}$\;
Drop $C_t \setminus K_t$ from the layer cache; \Return $K_t$\;
\caption{\method{} eviction at a single compression checkpoint.}
\label{alg:avokv-loop}
\end{algorithm}

Every quantity used at checkpoint~$t$ is already available in the decoded prefix. Read pressure uses only attention rows from queries in $O_t$, redundancy compares stored RoPE-transformed keys in $C_t$, and payload potential uses stored value norms. The method therefore changes the cache-selection rule without altering ordinary generation between compression events.

\textbf{Fixed configuration.} Compression is triggered every $128$ generated tokens. We use $W_{\mathrm{obs}}{=}8$, protect the first four cache entries as described above, and apply percentile-rank normalization. The protected tail is budget-dependent: $W_{\mathrm{delay}}{=}8$ for budgets at or above $768$ entries and $W_{\mathrm{delay}}{=}32$ for budgets at or below $512$, reflecting that tighter caches need more time to absorb late first reads. This single configuration was selected on small MATH-100/MATH-200 pilots before MATH-500 and reused unchanged across models, benchmarks, and budgets.

\section{Experiments}
\label{sec:experiments}

\subsection{Experimental Setup}
\label{sec:setup}

\paragraph{Models, tasks, and metrics.} We evaluate DeepSeek-R1-Distill-Llama-8B \citep{deepseekr12025} and Qwen3-14B \citep{qwen3technicalreport2025} on MATH-500 \citep{hendrycks2021math} and AIME-25 \citep{aime2025}. MATH-500 uses one sample per problem, whereas AIME-25 reports mean accuracy over $16$ samples per problem. Both models use sampled decoding with temperature~$0.6$ and top-$p\,0.95$, subject to a $32$K-token maximum length. All methods share the prompts, decoding loop, answer extractor, MATH-style ground-truth parser, and periodic compression checkpoints. Accuracy is our primary metric; we additionally record average output length, max-length hit count, and the maximum measured active KV length across compression events.

\paragraph{Baselines and matched-budget controls.} We compare against \fullkv{}, the uncompressed reference; \rkv{} \citep{rkv2025}, which uses the public attention\,$\times$\,redundancy score with $\lambda{=}0.1$, observation window $\alpha{=}8$, and compression interval~$128$; and \lazy{} \citep{lazyeviction2025}, which uses lagged attention-activity tracking with the same window and interval. Because \thinkv{} \citep{thinkv2025} had no public code release at the time of writing, we implement its published layer-wise, thought-adaptive mechanism using retention schedule $\{64,32,16,8,4\}$ and refresh interval~$128$ in the same eviction-only harness, and label it \thinkv{}~(impl.) throughout. All compressed methods are compared at matched measured active-KV budgets; AIME-25 uses a larger budget grid because its reasoning traces are longer. To isolate the eviction decision, retained entries remain in full precision, budget allocation is uniform across layers and heads, and no method receives a thought classifier, milestone detector, or serving-kernel modification.

\paragraph{\method{} configuration.} Compression is triggered every $128$ generated tokens. We use $W_{\mathrm{obs}}{=}8$, protect the first four cache entries, apply percentile-rank normalization, and set $(\lambda_p,\lambda_r,\lambda_b){=}(0.1,0.9,0.05)$. The protected delay is $W_{\mathrm{delay}}{=}8$ for budgets at or above $768$ entries and $W_{\mathrm{delay}}{=}32$ for budgets at or below $512$. This deterministic configuration was selected once on small MATH-100/MATH-200 pilots before MATH-500 and then reused unchanged across both models, both benchmarks, and every reported budget; we do not retune it per model or benchmark.

\begin{figure}[tbp]
\centering
\small
\includegraphics[width=0.99\textwidth]{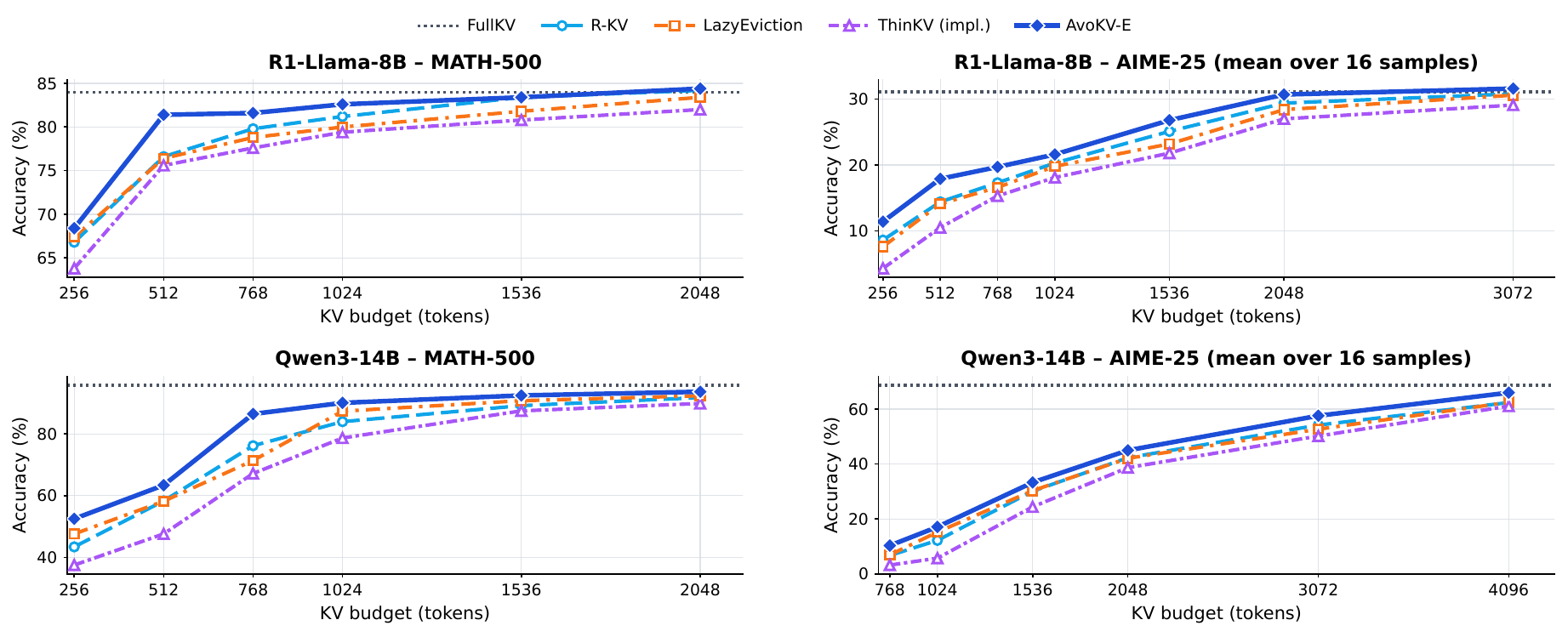}
\caption{Accuracy versus active KV budget for DeepSeek-R1-Distill-Llama-8B (top row) and Qwen3-14B (bottom row) on MATH-500 (left column) and AIME-25 (right column); AIME-25 reports mean accuracy over $16$ samples per problem. \fullkv{} is the dotted uncompressed reference. At matched budgets, \method{} matches or exceeds the redundancy-aware \rkv{}, recurrence-based \lazy{}, and thought-adaptive \thinkv{}~(impl.) baselines, with the largest gaps at the smallest budgets.}
\label{fig:budget-curve}
\end{figure}

\subsection{Accuracy and Termination under Matched Budgets}
\label{sec:results}

\paragraph{Accuracy--budget trade-off.} Figure~\ref{fig:budget-curve} plots accuracy against measured active-KV budget for both tasks and models. Each panel should be read horizontally: at a fixed memory budget, the higher curve represents the better keep set. \rkv{} and \lazy{} are the closest redundancy- and recurrence-based baselines, respectively, so their comparison with \method{} tests the benefit of adding payload-aware scoring. \method{} is best or tied-best throughout the shared budget range in all four panels. Its largest advantages occur at the tightest budgets, where the routing-only failure modes identified in \S\ref{sec:intro} are expected to be most damaging; as the budget grows, all compressed policies converge toward \fullkv{}.

\begin{table}[t]
\centering
\small
\setlength{\tabcolsep}{4pt}
\begin{tabular}{lccc}
\toprule
Method               & 256 & 512 & 1024 \\
\midrule
\fullkv{}            & --   & --   & --   \\
\rkv{}               & 105 & 78  & 60  \\
\lazy{}              & 112 & 85  & 65  \\
\thinkv{} (impl.)    & 119 & 92  & 70  \\
\method{}            & \textbf{102} & \textbf{55} & \textbf{50} \\
\midrule
\multicolumn{4}{l}{\emph{Headline accuracy on MATH-500 / R1-8B} (\%)} \\
\fullkv{}            & --     & --     & --     \\
\rkv{}               & 66.8   & 76.6   & 81.2   \\
\lazy{}              & 67.4   & 76.4   & 80.0   \\
\thinkv{} (impl.)    & 63.8   & 75.6   & 79.4   \\
\method{}            & \textbf{68.4} & \textbf{81.4} & \textbf{82.6} \\
\bottomrule
\end{tabular}
\caption{Max-length hit counts at the $32$K cap (out of $500$ MATH-500 problems) and headline accuracy at three representative KV budgets on DeepSeek-R1-Distill-Llama-8B. \fullkv{} terminates well within $32$K on essentially every problem. Among compressed methods, \method{} achieves both the fewest max-length hits and the highest accuracy at each budget.}
\label{tab:termination}
\end{table}

\paragraph{Consistency across evaluated settings.}
Although absolute accuracies and useful budget ranges differ across the four panels, the same budget-dependent pattern appears in each. \method{} separates most clearly from the competing eviction rules when capacity is scarce, then approaches them as the cache grows and fewer consequential choices are forced. The pattern also persists across both reasoning models and both benchmarks despite using the same configuration throughout. This consistency suggests that the improvement comes from the retention rule rather than from a model-, benchmark-, or budget-specific operating point.

\paragraph{Termination behavior.} The accuracy gains are accompanied by more stable completion behavior. The $32$K cap is long enough for \fullkv{} to terminate on essentially every prompt, so the remaining cap hits primarily reflect failures introduced by eviction. As Table~\ref{tab:termination} shows, \method{} obtains both the highest accuracy and the fewest max-length hits at budgets~$256$, $512$, and~$1024$. The separation is largest at budget~$512$, where \method{} produces $55$ cap hits compared with $78$ for the next-best baseline. Thus, preserving value payloads improves not only final-answer accuracy but also the model's ability to sustain and complete its reasoning trajectory.

\subsection{Why Does \method{} Work?}
\label{sec:analysis}

\paragraph{Counterfactual oracle ranking.} To isolate the score from downstream decoding, we evaluate several scoring rules against single-token oracle eviction-KL labels for $5{,}440$ aligned candidate KV entries from $30$ DeepSeek-R1 AIME-2024 traces. We report AUROC, average precision (AP), top-$k$ recall, false-safe rate, and Spearman correlation. This analysis complements the leave-one-out KL diagnostic in App.~\ref{app:motivation}.

\begin{table*}[t]
\centering
\small
\setlength{\tabcolsep}{8pt}
\renewcommand{\arraystretch}{1.15}
\begin{tabular}{lccccc}
\toprule
\textbf{Score} & \textbf{AUROC} & \textbf{AP} & \textbf{Top-$k$ recall} & \textbf{False-safe rate} & \textbf{Spearman} \\
\midrule
attention only          & 0.891 & 0.369 & 0.621 & 0.033 & 0.322 \\
\rkv{}-style proxy      & 0.894 & 0.372 & 0.629 & 0.026 & 0.303 \\
\thinkv{}-style proxy   & 0.892 & 0.350 & 0.610 & 0.022 & 0.303 \\
attention~$\times$~age  & 0.891 & 0.362 & 0.621 & 0.022 & 0.268 \\
\method{} (full score)  & \textbf{0.901} & \textbf{0.388} & \textbf{0.658} & \textbf{0.020} & \textbf{0.331} \\
\bottomrule
\end{tabular}
\caption{Oracle ranking quality against the single-token eviction-KL label on $5{,}440$ candidates. \method{}'s composite score has the strongest ranking metrics across the board while remaining within the same runtime budget as the proxy scores.}
\label{tab:oracle-rank}
\end{table*}

The full \method{} score is strongest on every ranking metric. The largest improvement appears in top-$k$ recall, which most directly measures how many high-damage entries the policy preserves. Because this comparison is independent of the decoding loop and uses the same runtime budget for all proxy scores, it links the downstream accuracy gains to the proposed payload-aware ranking rule.

The five metrics probe complementary aspects of the same ranking problem. AUROC and Spearman measure global ordering, while AP and top-$k$ recall emphasize the high-damage candidates that should remain near the top of the keep list. The false-safe rate measures the converse error: damaging entries assigned to the apparently safe-to-evict region. \method{} improves all five simultaneously, rather than improving one portion of the ranking at the expense of another. This is particularly relevant to online eviction, where mistakes near the keep--evict boundary become irreversible once an entry is removed.

\paragraph{Component ablations.}\label{sec:ablations} 
The largest ablation loss at every tested budget comes from removing key redundancy. This clarifies the role of payload awareness: \method{} does not replace routing evidence, but augments a strong routing backbone where that evidence is incomplete. Removing the payload term still lowers accuracy at every budget, consistent with the low-read/high-payload blindspot in Figure~\ref{fig:motivation}. Removing delayed eligibility also consistently hurts, as expected when recent states can be judged before later queries have had an opportunity to read them. Payload awareness and delayed eligibility therefore correct two different limitations of routing-only eviction: what evidence is considered and when that evidence becomes trustworthy.

The raw-score ablation shows that candidate-relative normalization is not merely a cosmetic rescaling. Read pressure, redundancy, and payload potential occupy different numerical ranges, so combining them directly changes their effective balance even when the nominal weights remain fixed. The degradation across all three budgets supports the scale-robust construction in Eq.~\ref{eq:normalizer}. Separately, the in-runner \rkv{} anchor closely tracks the official \rkv{} results, making a harness mismatch an unlikely explanation for the improvement of \method{}.

\begin{table}[H]
\centering
\small
\setlength{\tabcolsep}{4pt}
\begin{tabular}{lccc}
\toprule
Variant                              & 256 & 512 & 1024 \\
\midrule
\method{}                            & \textbf{68.4} & \textbf{81.4} & \textbf{82.6} \\
\quad $-$ payload ($\lambda_b{=}0$)  & 66.8 & 78.8 & 80.0 \\
\quad $-$ delay ($W_{\mathrm{delay}}{=}W_{\mathrm{obs}}$) & 66.4 & 79.6 & 81.0 \\
\quad $-$ rank norm. (raw)           & 64.2 & 76.8 & 78.7 \\
\quad $-$ redundancy ($\lambda_r{=}0$) & 60.5 & 73.4 & 76.4 \\
\midrule
\rkv{}-style anchor (in-runner)      & 66.6 & 76.4 & 80.2 \\
\rkv{} (official)                    & 66.8 & 76.6 & 81.2 \\
\bottomrule
\end{tabular}
\caption{Component decomposition of \method{} on MATH-500 with DeepSeek-R1-Distill-Llama-8B. Accuracy is reported at matched active-KV budgets. The in-runner \rkv{}-style anchor uses the same HuggingFace harness with the \rkv{} score and tracks the official \rkv{} row to within $\leq 1$ point at every budget.}
\label{tab:component-full}
\end{table}

\section{Conclusion}
This paper addresses KV-cache compression for long-output reasoning with \method{}, a training-free policy that separates delayed eligibility from payload-aware eviction risk. Across two reasoning models and two mathematical reasoning benchmarks, \method{} matches or exceeds the tested redundancy-aware, recurrence-based, and thought-adaptive baselines at matched active-KV budgets, with its largest gains under the tightest budgets. The oracle ranking and component analyses connect these gains to preserving high-damage value payloads, allowing late reads to materialize, and normalizing heterogeneous signals within the candidate set. More broadly, the results suggest that reasoning-cache eviction should preserve both retrievable keys and the value payloads that sustain the reasoning trajectory.

\section*{Limitations}
We evaluate two reasoning models (DeepSeek-R1-Distill-Llama-8B and Qwen3-14B) on two mathematical reasoning benchmarks (MATH-500 and AIME-25); we do not claim broad task coverage and have not tested code or multimodal reasoning. The prototype runs in HuggingFace generation rather than a serving kernel, so we report active KV budgets and decoding behavior, not packed-memory savings or end-to-end throughput. \method{} is an eviction policy; evaluating it alongside low-bit KV quantization and hybrid eviction--quantization allocation is a natural next step.

\bibliographystyle{plainnat}
\bibliography{references}

\clearpage
\appendix
\renewcommand{\theHtable}{\Alph{section}.\arabic{table}}
\renewcommand{\theHfigure}{\Alph{section}.\arabic{figure}}

\section{Worked Example}
\label{app:walkthrough}

We illustrate one compression checkpoint with $|C_t|=4$ eligible
candidate entries, using a single layer/head for brevity.
Table~\ref{tab:walkthrough} lists the raw features
$p_i,r_i,b_i$, their within-candidate percentile-rank normalized
values, and the resulting eviction-risk score $S_i$. The example is
constructed so that entries~1 and~3 are close under routing-only
evidence, but entry~3 carries a much larger value payload.

\begin{table}[ht]
\centering
\small
\setlength{\tabcolsep}{4pt}
\begin{tabular}{l cccc}
\toprule
$i$ & 1 & 2 & 3 & 4 \\
    & I & cook & dinner & myself \\
\midrule
\multicolumn{5}{@{}l}{\emph{Raw features}} \\
$p_i$ (read pressure)  & 0.04 & \textbf{0.81} & 0.02 & 0.05 \\
$r_i$ (key redundancy) & 0.55 & 0.41 & 0.55 & \textbf{0.92} \\
$b_i$ (value payload)  & 0.55 & 0.62 & \textbf{2.10} & 1.20 \\
\midrule
\multicolumn{5}{@{}l}{\emph{Percentile-rank normalized}} \\
$\hat p_i = N_t(p_i)$ & $1/3$ & $1$ & $0$ & $2/3$ \\
$\hat r_i = N_t(r_i)$ & $1/2$ & $0$ & $1/2$ & $1$ \\
$\hat b_i = N_t(b_i)$ & $0$ & $1/3$ & $1$ & $2/3$ \\
\midrule
\multicolumn{5}{@{}l}{\emph{Eviction-risk score}
\;($\lambda_p{=}0.1,\;\lambda_r{=}0.9,\;\lambda_b{=}0.05$)} \\
$S_i$ & $-0.42$ & $\mathbf{+0.12}$ & $\mathbf{-0.40}$ & $-0.80$ \\
\midrule
\multicolumn{5}{@{}l}{\emph{Routing-only score for comparison}
\;($\lambda_b{=}0$)} \\
$S_i^{\mathrm{route}}$ & $\mathbf{-0.42}$ & $\mathbf{+0.10}$ & $-0.45$ & $-0.83$ \\
\bottomrule
\end{tabular}%
\caption{Worked example at one compression checkpoint. The score is
$S_i = \lambda_p\,\hat p_i - \lambda_r\,\hat r_i + \lambda_b\,\hat b_i$.
Scores are rounded to two decimals. The tied redundancy values for
entries~1 and~3 are assigned the averaged percentile rank, following
the definition of $N_t$.}
\label{tab:walkthrough}
\end{table}

\paragraph{Reading the candidates.}
Entry~2 is read often and is not redundant, so both routing-only and
payload-aware scoring preserve it. Entry~4 has the most replaceable key,
so its high redundancy lowers its eviction-risk score despite moderate
read pressure and payload. Entries~1 and~3 are the informative pair:
entry~1 is a generic candidate with slightly higher read pressure,
whereas entry~3 is barely read but carries the largest value payload.

\paragraph{Selection under budget.}
With $B-|P_t|=2$, \method{} keeps the two eligible candidates with the
largest eviction-risk scores:
\[
  K_t \;=\; P_t \,\cup\, \operatorname{Top}_2\{S_i : i \in C_t\}
        \;=\; P_t \,\cup\, \{2,\,3\}.
\]
Thus, \method{} preserves both the strongly read entry and the low-read
but high-payload entry. If the payload term is removed, the routing-only
scores become
\[
  S^{\mathrm{route}} = (-0.42,\;0.10,\;-0.45,\;-0.83),
\]
so the top-2 set is instead $\{2,1\}$ and entry~3 is evicted. This
illustrates the attention-only blindspot from App.~\ref{app:motivation}:
a candidate can look slightly less useful under routing evidence while
still carrying a value payload whose removal is damaging.

\section{Extended Positioning}
\label{app:positioning}

Table~\ref{tab:related-positioning} summarizes how we position \method{} relative to nearby KV-cache methods. \method{} isolates the eviction decision and changes the score used to decide which active KV entries remain in cache; it does not allocate precision, classify thought regimes, or modify the serving kernel.

\begin{table*}[t]
\centering
\small
\renewcommand{\arraystretch}{1.35}
\setlength{\tabcolsep}{4pt}
\resizebox{\textwidth}{!}{%
\begin{tabular}{@{}>{\raggedright\arraybackslash}p{0.155\textwidth}>{\raggedright\arraybackslash}p{0.305\textwidth}>{\raggedright\arraybackslash}p{0.225\textwidth}>{\raggedright\arraybackslash}p{0.265\textwidth}@{}}
\toprule
\textbf{Family} & \textbf{Representative work} & \textbf{Main cache signal} & \textbf{Relation to \method{}} \\
\midrule
Attention retention &
H2O~\citep{h2o2023}, StreamingLLM~\citep{streamingllm2023}, Scissorhands~\citep{scissorhands2023}, TOVA~\citep{tova2024} &
Recent attention, heavy hitters, sinks, or persistence &
Also uses attention, but delays eligibility (delayed-read effect)
and adds value-payload potential (attention-only blindspot). \\
\midrule
Observation-window eviction &
SnapKV~\citep{snapkv2024}, NACL~\citep{nacl2024} &
Attention statistics over a short observation window &
Causal observation window applied \emph{during generated reasoning}
rather than only prompt-side selection. \\
\midrule
Budget allocation &
FastGen~\citep{fastgen2024}, PyramidKV~\citep{pyramidkv2024}, Ada-KV~\citep{adakv2024}, HeadKV~\citep{headkv2024}, Quest~\citep{quest2024}, DuoAttention~\citep{duoattention2024} &
Layer/head/query-aware allocation of limited cache capacity &
Allocation is held fixed; we study the token-level eviction score, so
this axis is orthogonal. \\
\midrule
Reasoning-cache compression &
RaaS~\citep{raas2025}, \rkv{}~\citep{rkv2025}, \lazy{}~\citep{lazyeviction2025}, \thinkv{}~\citep{thinkv2025}, LongFlow~\citep{longflow2026} &
Reasoning milestones, redundancy, lagged recurrence, thought regimes,
or current-query importance &
Closest baselines. Unlike \lazy{}, delay only defines eligibility while
payload-aware scoring drives the keep decision; unlike \thinkv{}, no
thought classifier or quantization. \\
\midrule
Quantization \& serving &
KIVI~\citep{kivi2024}, KVQuant~\citep{kvquant2024}, PM-KVQ~\citep{pmkvq2025}, TurboQuant~\citep{turboquant2026}, PagedAttention~\citep{vllm2023} &
Lower precision or more efficient cache storage &
Complementary; we disable quantization in our experiments to isolate
the eviction decision. \\
\bottomrule
\end{tabular}%
}
\caption{Positioning of \method{} among related KV-cache methods. Each row groups methods by their primary cache signal; the right-most column states what \method{} adds or holds fixed relative to that family.}
\label{tab:related-positioning}
\end{table*}

\section{Additional Reproducibility Details}
\label{app:implementation}

The evaluation protocol, baseline configurations, compression schedule, and fixed \method{} configuration are described in \S\ref{sec:setup}. Here we record the causal availability of the scoring signals, the pilot observations behind the fixed configuration, and the terms of the models and datasets used in the evaluation.

\paragraph{Causal scoring.} All three signals consume only information already in the prefix at checkpoint~$t$. The read-pressure feature averages attention rows from queries in $O_t$ over layers and heads; the redundancy feature compares RoPE-transformed keys within the candidate set $C_t$; the payload feature depends only on stored value norms. No future tokens, oracle labels, or auxiliary networks are used. This is what makes the score deployable inside ordinary autoregressive decoding.

\paragraph{Configuration-selection rationale.} The MATH-100/MATH-200 pilot pass yielded three observations that motivated the fixed choices in \S\ref{sec:setup}. First, the protected tail scales with the budget: at budget~$1024$, a short delay $W_{\mathrm{delay}}{=}8$ is optimal, whereas at budget~$512$, a longer delay $W_{\mathrm{delay}}{=}32$ is consistently more stable because tighter budgets evict more aggressively and need a wider window to absorb late first reads. Second, the payload weight is small but non-zero: setting $\lambda_b{=}0$ costs several points across budgets, while pushing $\lambda_b$ much beyond $\sim 0.05$ degrades behavior. The payload term is intended to break ties among similarly routed candidates and rescue the high-payload states identified by the attention-only blindspot, not to dominate the score; the relatively large redundancy weight $\lambda_r{=}0.9$ keeps the overall score close to the strong \rkv{} family. Third, percentile-rank and age-bucket-rank normalization are interchangeable in the pilots, whereas the scale-sensitive normalizers we tried (raw, $z$-score, min--max, max-absolute, and softmax) collapse the score toward attention-only behavior and forfeit $3$--$11$\,points depending on budget. Under the same harness, we also compared the full \method{} score against \rkv{}+payload, a recency-weighted variant, and an \rkv{}-style anchor; the composite score was consistently best at both pilot budgets.

\paragraph{Artifacts and terms.} We use publicly available pretrained models and math benchmarks only for research evaluation. DeepSeek-R1-Distill-Llama-8B is released by DeepSeek-AI under the MIT license, with the upstream Llama 3.1 license noted by the model provider; Qwen3-14B is released under Apache-2.0; and the Hugging Face version of the Hendrycks MATH dataset is listed under the MIT license. AIME-25 problems are credited to the Mathematical Association of America / public problem sources and are used only for benchmark evaluation. We do not redistribute model weights or benchmark contents.

\section{Empirical foundations of Figure~\ref{fig:motivation}}
\label{app:motivation}

Two intuitions underlie \method{}: that attention magnitude alone is a lossy summary of a token's importance because it ignores the size of what the token would actually contribute if read, and that a single recent observation window is too short a horizon to reliably classify reasoning tokens as useful or stale. Figure~\ref{fig:motivation} presents these ideas in takeaway form; this appendix gives the detailed construction behind the figure, including the candidate pool, the oracle high-damage definition, the attention/value-norm bins in panel~(a), the strong-read lag statistic in panel~(b), and the significance tests behind the reported blindspot.

\paragraph{Candidate pool.} We pool $n{=}892$ candidate KV entries from five MATH/AIME traces generated by DeepSeek-R1-Distill-Llama-8B. A candidate is a (trace, position) pair whose remaining horizon supports the longest measurement window required below ($W_{\text{fut}}{=}8$ steps for oracle damage, up to $32$ steps for the lag analysis). The candidates are not subsampled within a trace, so the marginals on the panel-(a) heatmap reflect the actual distribution of attention and value-norm ranks under decoded reasoning, not an attention-stratified subset. Each candidate is described by the two within-trace percentile ranks introduced in \S\ref{sec:method}: recent attention $N_t(p_i)$, aggregated over the last $W_{\text{obs}}$ decode steps (Eq.~\ref{eq:read-pressure}); and value-norm $N_t(b_i){=}\|v_i\|_2$, the $\ell_2$ magnitude of the candidate's stored value vector — the $v$ from QKV that would be returned via attention if the entry were retained (Eq.~\ref{eq:payload}).

\paragraph{Oracle damage and the attention-only blindspot (panel~a).} For each candidate we want a damage signal that does not assume any specific eviction policy. The leave-one-out oracle achieves this directly: we replay the next $W_{\text{fut}}{=}8$ decode steps twice from the same context, once with the full cache and once with candidate $i$'s KV entry removed, and measure the average KL between the two next-token distributions,
\begin{equation}
  D_i^{\text{KL}}
  \;=\;
  \frac{1}{W_{\text{fut}}}\sum_{\tau=1}^{W_{\text{fut}}}
  \mathrm{KL}\!\left(p_\tau^{\text{full}}\,\Big\|\,p_\tau^{(-i)}\right).
  \label{eq:oracle-damage}
\end{equation}
$D_i^{\text{KL}}$ has very different absolute scales across traces, so we threshold within trace: a candidate is \emph{high-damage} if its oracle damage falls in the top decile of its host trace. This construction guarantees a $\approx 10\%$ baseline across the full candidate pool, against which any cell in the (attention, value-norm) grid can be compared.

Out of the $887$ candidates for which the future window is complete, the blindspot cell $\{N_t(p_i){\leq}0.2,\allowbreak\, N_t(b_i){\geq}0.8\}$ contains $10/43{=}23\%$ high-damage states (Wilson $95\%$ CI $[13\%, 38\%]$), against $81/844{=}9.6\%$ over the remaining $22$ cells. The enrichment is $2.4{\times}$ and one-sided Fisher's exact gives $p{=}0.009$. Two further cells survive a per-cell one-sided binomial test against the $9.6\%$ baseline, both in the high-attention column ($N_t(p_i){\geq}0.8$): $8/41{=}20\%$ and $7/35{=}20\%$, both $p{<}0.05$. These two cells act as a positive control — they confirm that the analysis recovers the obvious effect (high-attention tokens are damaging to remove) — which makes the blindspot cell stand out all the more strongly: it is the only significantly elevated cell \emph{outside} the high-attention column. The marginal distributions along each axis are individually flat, so neither attention nor value-norm alone discriminates damaging from safe candidates; both features are needed.

\paragraph{Delayed first reads and the case for delayed observation (panel~b).} The blindspot exists because attention-only scoring is a snapshot at a single moment. Whether that snapshot is informative depends on how quickly reasoning tokens are read after generation. We define a \emph{strong read} of candidate $i$ as a future query position $\tau$ whose attention $\alpha_{\tau,i}$ reaches at least $0.85$ of $i$'s per-trace peak attention, and the \emph{first-strong-read lag} of $i$ as the smallest such $\tau-i$. Lags are right-censored at $32$ decode steps, and we restrict the analysis to the $n{=}542$ candidates whose remaining horizon supports that cap so the distribution is not artificially truncated.

The diagnostic quantity is the cumulative read probability $F(t){=}P(\text{lag}{\leq}t)$, the fraction of candidates that have been strongly read by lag $t$. Splitting candidates by attention quartile ($N_t(QK^{\!\top}){\leq}0.4$ vs.\ $N_t(QK^{\!\top}){\geq}0.6$; $n{=}218$ each), $F(4){=}51\%$ vs.\ $62\%$ and $F(16){=}56\%$ vs.\ $72\%$; the high-attention curve dominates throughout and the gap to the low-attention curve \emph{widens} with $t$ (one-sided Kolmogorov--Smirnov $p{=}0.010$). The plateau is informative on its own: $15\%$ of low-attention candidates are never strongly read inside the $32$-step horizon at all. Reading this from the eviction side, an attention-only policy that decides at small $t$ is precisely the policy that conflates ``not yet read'' with ``not useful'', and the worst-case lag is concentrated on exactly the low-attention candidates the policy is most eager to discard. Waiting more decode steps before the decision --- the delayed causal observation construction in \S\ref{sec:method} --- allows these late reads to materialize, costing only bookkeeping but turning the gap in panel~(b) into a reliable keep/evict signal.

Together, panels (a) and (b) make a complementary case: the snapshot attention used by prior eviction methods is structurally blind to a high-damage subpopulation (panel~a), and it is unreliable at the short lags where those methods operate (panel~b). \method{} addresses both by combining a payload-aware score with delayed observation.

\end{document}